\documentclass[a4paper]{jpconf}
\usepackage{graphicx}
\usepackage{amsmath}
\begin{document}
\title{Physics-informed machine learning for moving load problems}

\author{Taniya Kapoor, Hongrui Wang, Alfredo Núñez, and Rolf Dollevoet}

\address{Section of Railway Engineering, Department of Engineering Structures,\\
Faculty of Civil Engineering and Geosciences, Delft University of Technology, The Netherlands}

\ead{t.kapoor@tudelft.nl}
% % h.wang-8, a.a.nunezvicencio, r.p.b.j.dollevoet\

\begin{abstract}
This paper presents a new approach to simulate forward and inverse problems of moving loads using physics-informed machine learning (PIML). Physics-informed neural networks (PINNs) utilize the underlying physics of moving load problems and aim to predict the deflection of beams and the magnitude of the loads. The mathematical representation of the moving load considered in this work involves a Dirac delta function \cite{yang2004vehicle}, to capture the effect of the load moving across the structure. Approximating the Dirac delta function with PINNs is challenging because of its instantaneous change of output at a single point, causing difficulty in the convergence of the loss function. We propose to approximate the Dirac delta function with a Gaussian function. The incorporated Gaussian function physical equations are used in the physics-informed neural architecture to simulate beam deflections and to predict the magnitude of the load. Numerical results show that PIML is an effective method for simulating the forward and inverse problems for the considered model of a moving load.
\end{abstract}

\section{Introduction}
Moving load problems are ubiquitous in engineering, where the structure is subjected to a moving point mass source. Examples of moving loads include a vehicle moving on a bridge, a train moving on railway tracks, and a pantograph moving on a catenary, among others. The moving point mass source induces forces on the structure, which subsequently causes the deformation and displacement of structures. One physical quantity of interest for engineers is the deflection of the structure. Over time, estimation is required for asset management, including the stages of design, operation, maintenance and replacement of the structure. In practice, numerical studies are carried out to compute the deflection profiles. However, in the case of transportation infrastructures such as railways, where the track infrastructures are hundreds of kilometers long and the deflection of the tracks can be in scales of centimeters or even lower, obtaining the complete deflection profiles of the railway tracks and catenaries is challenging. 

Mathematically, moving loads over the beam-like structures can be modeled as partial differential equations (PDEs), including a Dirac delta function modeling the moving point mass source. Extensive studies have been carried out to model and simulate the moving load problems using PDEs including but not limited to  \cite{ouyang2011moving, fuaruaguau2022dynamic, yang2020novel, shamalta2003analytical,  metrikine2006dynamic, mazilu2017dynamics, mazilu2017interaction, bao2022impact}. These PDEs could be simulated to obtain complete deflection profiles at all space-time locations. In recent years, considerable interest has shifted towards employing deep learning-based methods to simulate and model differential equations. Traditionally deep learning-based methods require a substantial amount of data; however, data collection under all possible scenarios of different realizations of stochasticities is not possible. For some scenarios, limited experimental data could be available, but the generalization to other physical parameters with no data measured is not guaranteed with traditional deep learning-based methods. One possible way to mitigate this challenge is to leverage the physical equations directly in the neural network loss function. Recently, a deep neural network-based architecture termed physics-informed neural networks (PINNs) has been proposed to solve physical equations that use the PDEs in the loss function along with initial and boundary conditions and does not need any additional experimental data to simulate the deflection profiles of the structures \cite{raissi2019physics}. 

%The latest research on PINN has shown significant progress in solving systems of physical equations, including nonlinear engineering systems. However, there are several limitations of PINNs, including their inability to tackle shock-type behaviour because of spectral bias and non-convex landscape of optimizer\cite{krishnapriyan2021characterizing,cuomo2022scientific}. A non-convex landscape refers to a situation where the function being optimized has multiple local optima. 

Previously, simply supported Euler-Bernoulli and Timoshenko beam PDEs have been simulated using PINNs for sinusoidal force function \cite{kapoor2023physics, bazmara2023physics}. The Dirac delta function in moving load PDE models moving point mass, which poses a challenge in the PINNs training process because of the instant discontinuous behaviour of the Dirac delta function. We aim to tackle this challenge and accurately simulate the moving load problem by incorporating a Gaussian approximated PDE in the loss function. This approximation results in a smooth estimation of the Dirac delta function, mitigating the challenge for the  PINN architecture.

This paper presents numerical experiments for forward and inverse problems of the moving load PDEs. Forward problems refer to solving the moving load PDE to obtain deflection profiles of the structure. Inverse problems refer to estimating the unknown physical parameters of the PDE provided some data for the deflection profiles at certain locations. The main contributions of this paper are as follows,
\begin{itemize}
\item  To our best knowledge, this is the first work which simulates the moving load problems through physics-informed machine learning.
\item From the machine learning perspective, this work proposes an approach to tackle the Dirac delta function for PINN training approaches. We approximate the Dirac delta function by a smooth Gaussian function to make possible the PINN training. We show the approach in the moving load PDE.
\item For the forward problem results show that PIML predicts the deflection of the beam under different magnitudes of moving loads. 
\item Additionally, for the inverse problem, PIML predicts the different magnitudes of the moving loads effectively. Traditional approaches for solving inverse problems require repeated solutions of the forward problem, which are computationally expensive. However, PIML for the inverse problem is computationally less expensive because they seamlessly incorporate prior knowledge and available data.
\end{itemize}

The rest of the paper is structured as follows: Section 2 presents the mathematical model for a moving load problem. Section 3 introduces PINNs for simulating the moving load problem. Section 4 presents the approximation of the Dirac delta function using a Gaussian function resulting in the PIML methodology. In addition, Section 4 also contains the numerical results for a forward problem and an inverse problem of a moving load on a simply supported beam. Finally, Section 5 concludes the paper and outlines potential future works.

\section{Mathematical model}

The mathematical model for a single load moving over a simply supported beam is based on several assumptions described in \cite{yang2004vehicle} as shown in Figure 1. The beam is assumed to have a constant cross-sectional area and homogeneous composition consistent with the Euler-Bernoulli hypothesis. The model assumes that only one vehicle can traverse the bridge at any given time and that the gravitational effect of the vehicle is the primary consideration. At the same time, its inertia is negligible compared to the bridge. The vehicle is assumed to move at a constant velocity, $v$, and the beam damping is modeled according to the Rayleigh method. Additionally, the beam is assumed to be at rest before the vehicle begins to move, and the surface roughness of the bridge is not taken into account. The mathematical model is defined as follows \cite{yang2004vehicle}:

\begin{figure}
\centering
\includegraphics[width=35pc]
{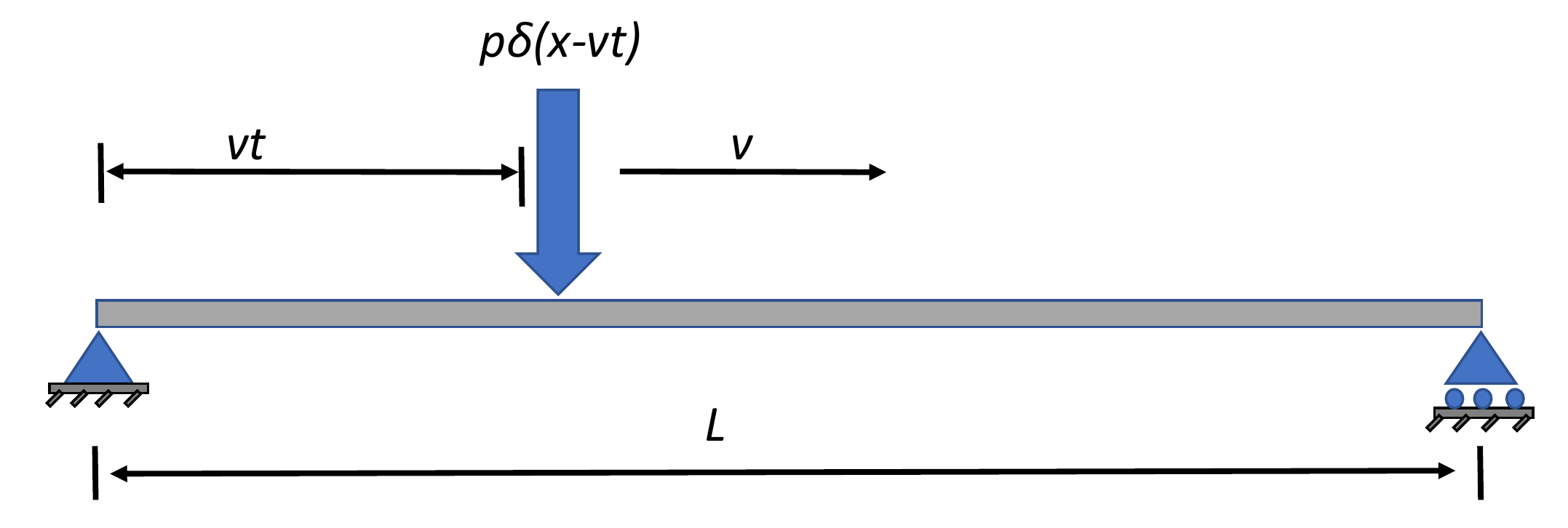}\hspace{2pc}%
\caption{\label{label1}A point mass source of load $p$ moving on a simply supported beam of length $L$ with velocity $v$.}
%\end{minipage}
\end{figure}

\begin{equation}
mu_{tt} + c_\mathrm{e}u_{t} + c_\mathrm{i}u_{txxxx} + EIu_{xxxx} = p\delta(x-vt), \quad 0 \leq vt \leq L
\label{eq1}
\end{equation}
where, $\delta$ is the Dirac delta function. For the simply supported beam, the boundary conditions are as follows,
\begin{equation}
u(0, t) = 0, \quad u(L, t) = 0, \quad u_{xx}(0, t) = 0, \quad u_{xx}(L, t) = 0   ; 
\end{equation}
Initially, the beam is assumed to be at rest; hence the initial conditions are as follows,
\begin{equation}
u(x, 0) = 0, \quad u_{t}(x, 0) = 0
\end{equation}

The deflection of the beam is denoted by $u(x, t)$, and $L$ represents the beam length. Notations m, $c_\mathrm{e}$, $c_\mathrm{i}$, $E$, $I$, $p$ and $v$ represent mass per unit length, external damping coefficient, internal damping coefficient, modulus of elasticity, the moment of inertia of the beam, magnitude of load and speed of load, respectively. Considering the eﬀect of damping over the beam to be small, we neglect the eﬀect of damping ($c_\mathrm{e}, c_\mathrm{i} = 0$) and the ground truth of the deflection $u(x, t)$ of the beam is given in \cite{yang2004vehicle} as,

\begin{equation}
u(x, t) =  \frac{2pL^3}{EI\pi^4}\sum_{n = 1}^{\infty}\frac{1}{n^4}\mathrm{\sin}\left(\frac{nx\pi}{L}\right)\left(\frac{\mathrm{\sin}\left(\Omega_{n}t\right) - S_{n}\mathrm{\sin}\left(\omega_nt\right)}{1 - S_n^2}\right)
\end{equation}
where, 

\begin{equation}
\Omega_n = \frac{nv\pi}{L},  \quad \omega_n = \frac{n^2\pi^2}{L^2}\sqrt\frac{EI}{m}, \quad S_n = \frac{\Omega_{n}}{\omega_n}
\end{equation}
 
%The calculation of the bending moment and shear force involves taking the second derivative of $u$ with respect to $x$, represented by $M = -EIu_\mathrm{xx}$, and the third derivative of $u$ with respect to $x$, represented by $V = EIu_\mathrm{xxx}$.

\section{Physics-informed neural networks}
PINNs are based on deep neural networks (DNNs) architecture. In supervised training of the DNNs, one needs input and output training data. More precisely, it requires output for every input location. For some problems, collecting data for all possible locations can be challenging. To mitigate the challenge of scarce data, prior knowledge of mathematical models or equations can be utilized to avoid the need for output at every location. For solving the forward problem, PINNs only require output data at initial and boundary points as proposed in \cite{raissi2019physics,karniadakis2021physics}. For interior points, prior knowledge of PDEs is utilized, and the desired derivatives of physical equations are calculated using the autograd feature of the neural network.

Since the PDE is ill-posed for the inverse problem, we need to provide additional data at a certain location or time to predict the physical parameters. To simulate the deflection profile of a simply supported beam with a moving point mass source, as shown in Figure 1 using PINN, the following steps are followed:

\begin{itemize}
\item The inputs are defined for the whole space-time domain $D \times T$ where $D = \Omega_\mathrm{int} \cup \Omega_\mathrm{b} \cup \Omega_\mathrm{0}$ and $T = \Gamma_\mathrm{0}  \cup   \Gamma_\mathrm{b} \cup \Gamma_\mathrm{int}$. The initial, boundary and interior training points are defined as  $(x_\mathrm{in}, t_\mathrm{in}) \in \Omega_0 \times \Gamma_0$; $(x_\mathrm{b}, t_\mathrm{b}) \in \Omega_\mathrm{b} \times \Gamma_\mathrm{b}$ and $(x_\mathrm{int}, t_\mathrm{int}) \in \Omega_\mathrm{int} \times \Gamma_\mathrm{int}$ respectively. For the well-posedness of the physical equation, we provide initial condition $u(x_\mathrm{in}, t_\mathrm{in})$, and boundary conditions $u(x_\mathrm{b}, t_\mathrm{b})$  as output data for training the PINN architecture. The neural network is trained using the physical equation with initial and boundary conditions.

\item The number of training points $N$ consists of $N_\mathrm{int}$, $N_\mathrm{in}$ and $N_{b}$ that are total interior training points, initial training and boundary training points. Training involves  minimizing the loss function using an optimizer, which comprises three terms as follows:
\begin{enumerate}

\item \textbf{PDE loss}: This quantity is determined by taking the derivatives of the predicted solution using automatic differentiation and minimization process of neural networks, ensuring that it satisfies the physical equation of moving loads. The loss term without damping is given as:
\begin{equation*}
L_\mathrm{PDE} = \sum_{i=1}^{\mathrm{N}_\mathrm{int}} |mu^{*}_{tt}(x_\mathrm{int}^i,t_\mathrm{int}^i) + EIu^{*}_{xxxx}(x_\mathrm{int}^i,t_\mathrm{int}^i) - p\delta(x_\mathrm{int}^i-vt_\mathrm{int}^i)|^2
\end{equation*}
	
 %$L_\mathrm{PDE}$ = $\sum_{i} |(u_{tt}(x_{int},t_{int}) + u_{xxxx}(x_{int},t_{int}) - p\delta(x-vt)|^2$

where $\delta(x_\mathrm{int},t_\mathrm{int})$ is a Dirac delta function which models the moving load in the PDE. 
\item \textbf{Boundary loss}: This quantity measures the deviation of the predicted solution and ground truth on boundary points. The PDE is well-posed. Therefore we have the ground truth for the boundary points. This term is calculated as the sum of the squared differences between the predicted and actual solution at the boundary training points. The loss term from boundary conditions is given as follows:
	\begin{equation*}
	    L_\mathrm{BC} = \sum_{i = 1}^\mathrm{N_{b}} |u^{*}(x_\mathrm{b}^i,t_\mathrm{b}^i) - u(x_\mathrm{b}^i,t_\mathrm{b}^i)|^2
	\end{equation*}
 
\begin{figure}
\centering
\includegraphics[width=20pc]
{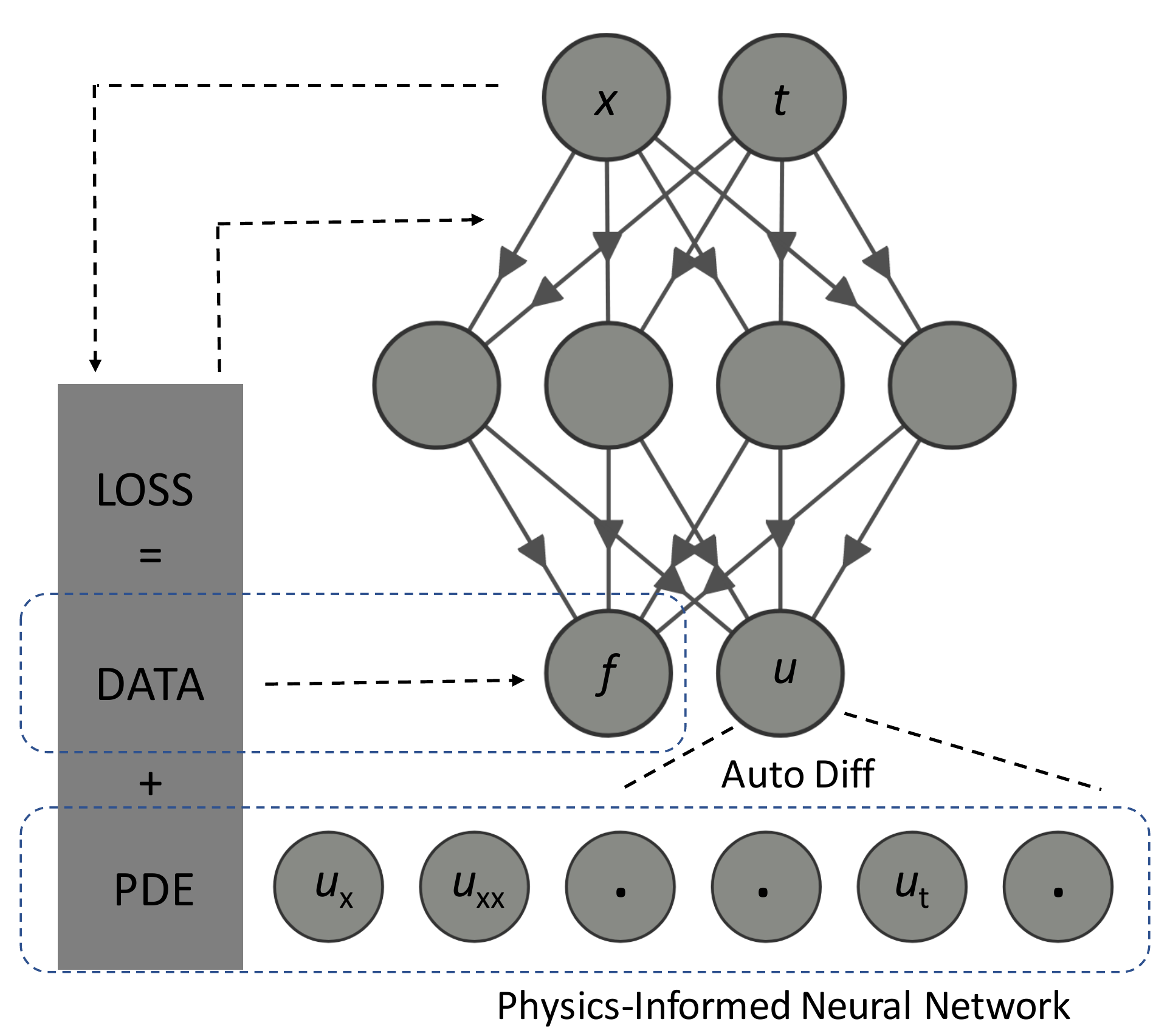}\hspace{2pc}%
\caption{\label{label2}PINN architecture for moving loads: for forward problems, the loss function comprises the PDEs, and data consist of the boundary and initial conditions. For the inverse problems, ill-posed PDEs are supplemented with additional data at certain locations.}
%\end{minipage}
\end{figure}

where $u(x_{b},t_{b})$ is a boundary condition function, and $u^*(x_{b},t_{b})$ is the predicted solution at the boundary point $(x_{b}, t_{b})$.
\item \textbf{Initial loss} This quantity measures the deviation of the predicted solution and ground truth on initial training points. The PDE is well-posed. Therefore we have the ground truth for the initial points. This term is calculated as the sum of the squared differences between the predicted and actual solution on the initial training points. The loss term from initial conditions can be given as follows:
 \begin{equation*}
     L_\mathrm{IC} = \sum_{i = 1}^{N_\mathrm{in}} |u^{*}(x_\mathrm{in}^i,t_\mathrm{in}^i) - u(x_\mathrm{in}^i,t_\mathrm{in}^i)|^2
 \end{equation*}

where $u(x_\mathrm{in},t_\mathrm{in})$ is a initial condition function, and $u^*(x_\mathrm{in},t_\mathrm{in})$ is the predicted solution at the initial point $(x_\mathrm{in}, t_\mathrm{in})$.

    \end{enumerate}

The total loss is calculated as the sum of three loss terms from the PDE, boundary, and initial losses:
\begin{equation*}
    Loss = \lambda_\mathrm{1}L_\mathrm{PDE} + \lambda_\mathrm{2}L_\mathrm{IC} + \lambda_\mathrm{3}L_\mathrm{BC}
\end{equation*}
\end{itemize}

Here, $\lambda_1, \lambda_2$ and $\lambda_3$ are weights in the loss function. These weights could be chosen on a prior assumption or trained adaptively. As shown in Figure 2, for the case of a forward problem, the only data we have is for initial and boundary points. 

For the inverse problem, PDE is ill-posed, which implies either the initial condition, boundary condition, or parameter of the physical equation is missing. For this case, we need to provide additional data at certain points over the domain, as shown in Figure 1. The loss function for the inverse problem is defined as follows: 

\begin{equation*}
    Loss = \lambda_\mathrm{1}L_\mathrm{PDE} + \lambda_\mathrm{2}L_\mathrm{IC} + \lambda_\mathrm{3}L_\mathrm{BC} + L_\mathrm{data}
\end{equation*}
where $L_\mathrm{data}$ is defined as;

\begin{equation*}
	    L_\mathrm{data} = \sum_{i = 1}^\mathrm{N_\mathrm{data}} |u^{*}(x_\mathrm{data}^i,t_\mathrm{data}^i) - u(x_\mathrm{data}^i,t_\mathrm{data}^i)|^2
	\end{equation*}
where $N_\mathrm{data}$ is the total number of additional data points at which the beam deflections on certain locations $(x_\mathrm{data},t_\mathrm{data})$ are known. 

\section{Gaussian approximation and numerical results}
This section presents the experiments for forward and inverse problems of moving loads. For the forward problem, the deflection of the beam is the quantity of interest where data is only provided for initial and boundary points. For the inverse problems, the quantity of interest is the magnitude of load $p$. 

\subsection{PINNs for the forward problem of moving load with the Dirac delta function}

The aim is to simulate the deflection of a simply supported beam for different magnitudes of loads. We consider the nondimensionalized form \cite{kapoor2023physics, kapoor2022predicting, kissas2020machine} of simple supported beam~\eqref{eq1}. The considered neural network architecture for the moving load problems are $1$ hidden layers, $20$ neurons, $\lambda_1 = 1$, $\lambda_2, \lambda_3 = 10$. Tanh activation function was chosen. The neural network was trained for 5000 epochs with only 1600 total training points. The Dirac delta function in equation~\eqref{eq1} is define as:
\begin{equation}
    \delta(x - t ) = \Bigg\{ \begin{array}{ll}
      1 & x - t  = 0 \\
      0 & x -t \neq 0 \\
\end{array}
\end{equation}

The choosen physical parameters in equation~\eqref{eq1} are  $v = 1$, $m = 1$, $L = \pi$,  $p = 1$, $E, I, \rho = 1$ and $x \in [0, \pi], t \in [0, \pi/2]$. As shown in Figure 3, results indicate that the PINN method cannot approximate a solution. The reason for this failure is the Dirac delta function on the right-hand side of the physical model of the moving loads. Results show that the vanilla PINNs fail to approximate the deflection of the beam effectively due to the instantaneous change at a single point represented by the Dirac delta function in the equation. This instantaneous change poses a challenge for neural network training. We propose approximating the Dirac delta function to address this issue with a smoother function. The idea is to achieve a smoother training process for the PINN architecture that models the same behaviour as the Dirac delta.

\begin{figure}[ht]
\centering
\includegraphics[width=20pc]{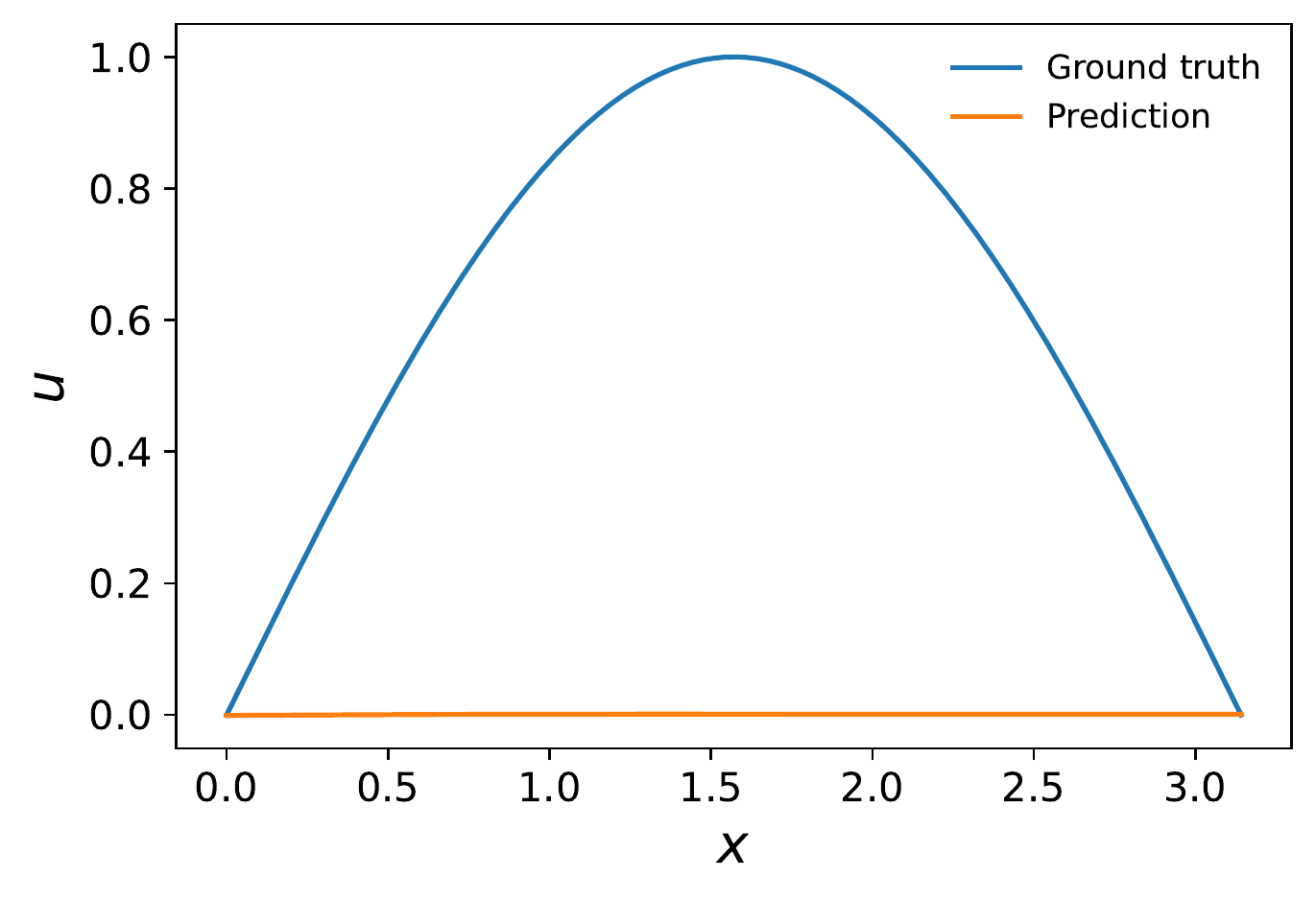}\hspace{2pc}%
\caption{\label{label3}PINN prediction for the moving load at $t=\pi/2$ with the Dirac delta function.}
\end{figure}
\begin{figure}[ht]
\begin{minipage}{18pc}
\includegraphics[width=18pc]{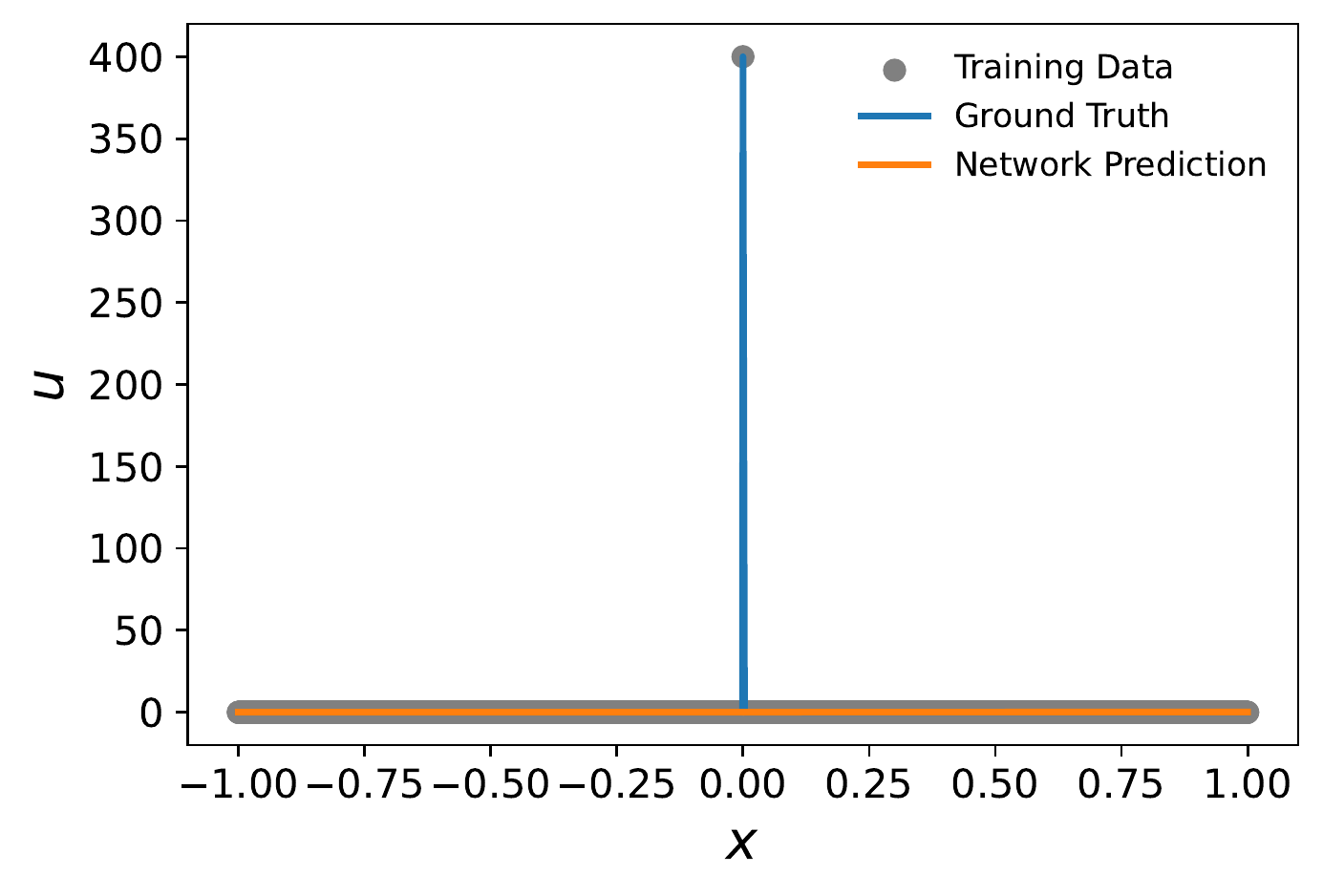}
\caption{\label{label4}Approximating the Dirac delta function using DNN.}
\end{minipage}\hspace{2pc}%
\begin{minipage}{18pc}
\includegraphics[width=18pc]{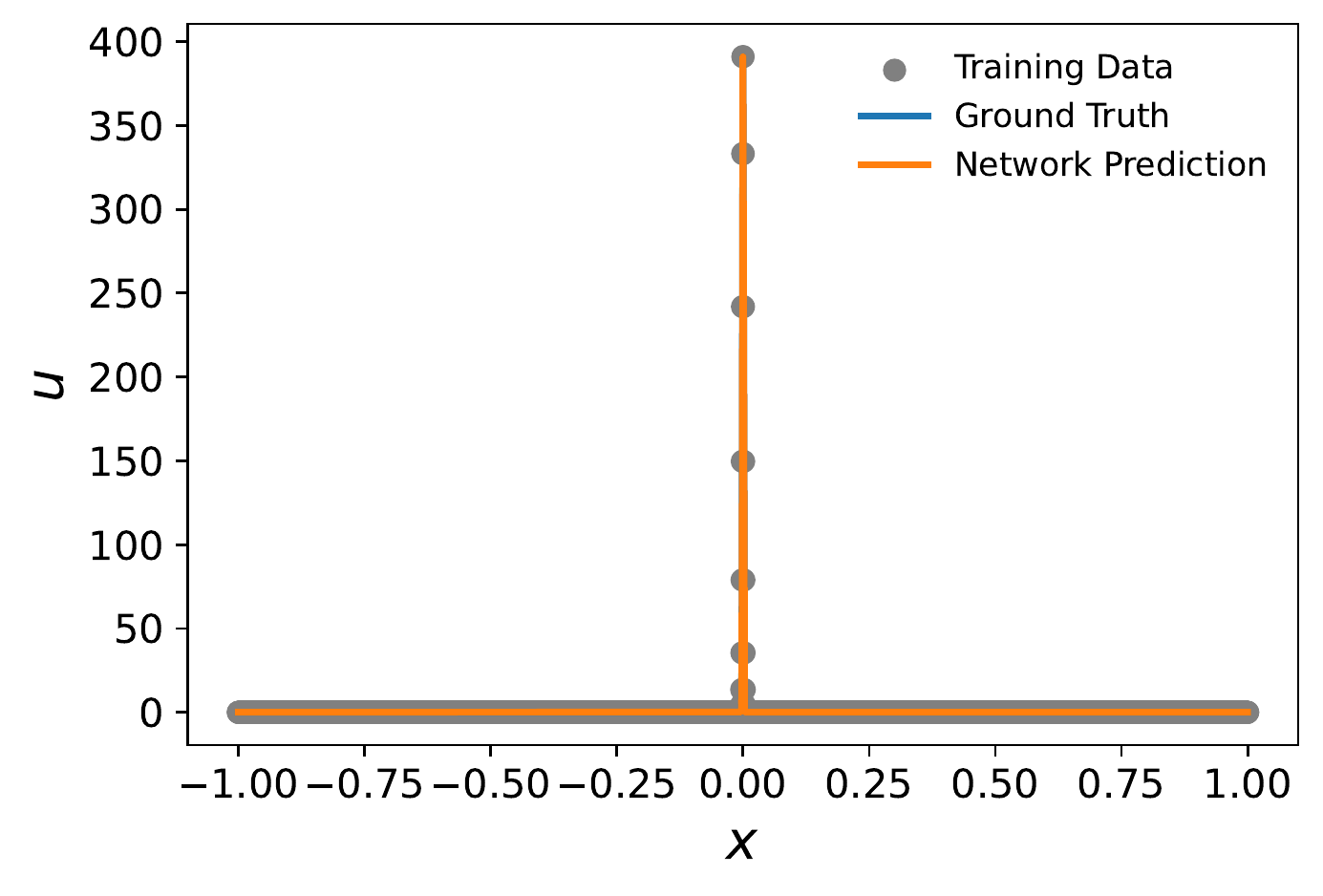}
\caption{\label{label5}Approximation of the Dirac delta with Gaussian function using DNN.}
\end{minipage} 
\end{figure}

\subsection{Gaussian approximation of the Dirac delta function}
The previous results indicate that the Dirac delta function impedes the convergence of the PINN architecture during training. This study aims to address this issue by approximating the Dirac delta function. First, we approximate the Dirac delta function using DNN and observe that the neural network training can be challenging because of sudden changes only at one location and the availability of only one training point, as shown in Figure 4.

To overcome this challenge, we propose approximating the Dirac delta function using a smooth function with a Gaussian function:
\begin{equation*}
    \delta(x) = \frac{1}{\sigma \sqrt{2\pi}}  e^{\left(\frac{\left(x- \mu\right)^2}{2 \sigma^2}\right)}
\end{equation*}
%\begin{equation*}
 %   \delta(x - t ) = \frac{1}{\sigma \sqrt{2\pi}}  e^{\left(\frac{\left(x-t - \mu\right)^2}{2 \sigma^2}\right)}
%\end{equation*}

The mean $\mu$ and variance  $\sigma$ are choosen to be $0$ and $0.001$. DNN is trained with $4$ hidden layers, $50$ neurons, tanh activation function and ADAM optimizer. The training process of the neural network converges with a relative error percentage of only $0.002\%$ as shown in Figure 5. where the relative error percent is defined as:
\begin{equation}
\begin{aligned}
\mathcal{R} = \frac{||u^{*} - u ||_{2}}{||u||_{2}} \times 100
\end{aligned}
\end{equation}

Hence, DNNs can effectively approximate the Gaussian function and provide a better alternative to approximating the Dirac delta function in the PINN architecture. 

\subsection{PIML for the forward problem of moving load with Gaussian function}
We observed that a Gaussian function approximates the Dirac delta function well, using DNN as shown in Figure 5 and proposed in \cite{huang2021solving}. Motivated by this example, we substitute the Dirac delta function with a Gaussian function in equation~\eqref{eq1} in the moving load problem. The mean and variance for the Gaussian function are chosen to be $0$ and $\frac{1}{\sqrt{2\pi}}$ We incorporated the new physical equation into the PINN architecture and trained the neural network using $1600$ total training points, with $1$ hidden layers, $20$ neurons, weights $\lambda_1, \lambda_3 = 10$ and $\lambda_2 = 1$. A tanh activation function is utilized for $5000$ epochs. The results show that the PIML architecture is trained well and accurately converged to a solution for various loading conditions, as demonstrated in Figure 6. The relative error percentage $R$ for different loads is less than 1$\%$, as indicated in Table $1$, demonstrating that the PIML effectively predicts deflection under different magnitudes of loads.
\begin{figure}[ht]
\centering
\includegraphics[width=25pc]{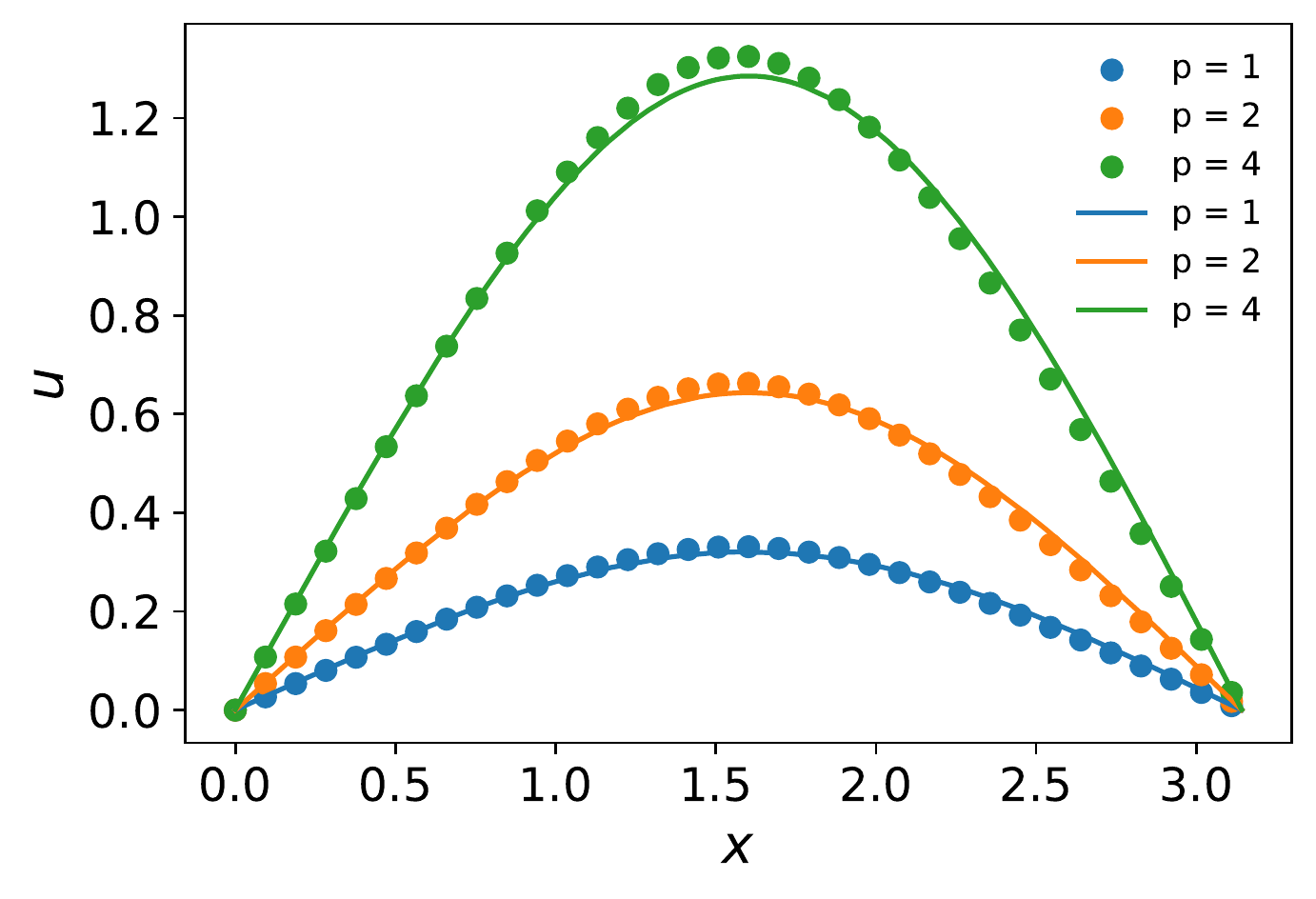}\hspace{2pc}%
\caption{\label{label6}Deflection of the beam at final time $(t = \pi/2)$, $p$ is the magnitude of the load in equation 1. Scatter plots represent the ground truth, and the solid lines represent the PIML predictions.}
\end{figure}
\begin{figure}[ht]
\begin{minipage}{18pc}
\includegraphics[width=18pc]{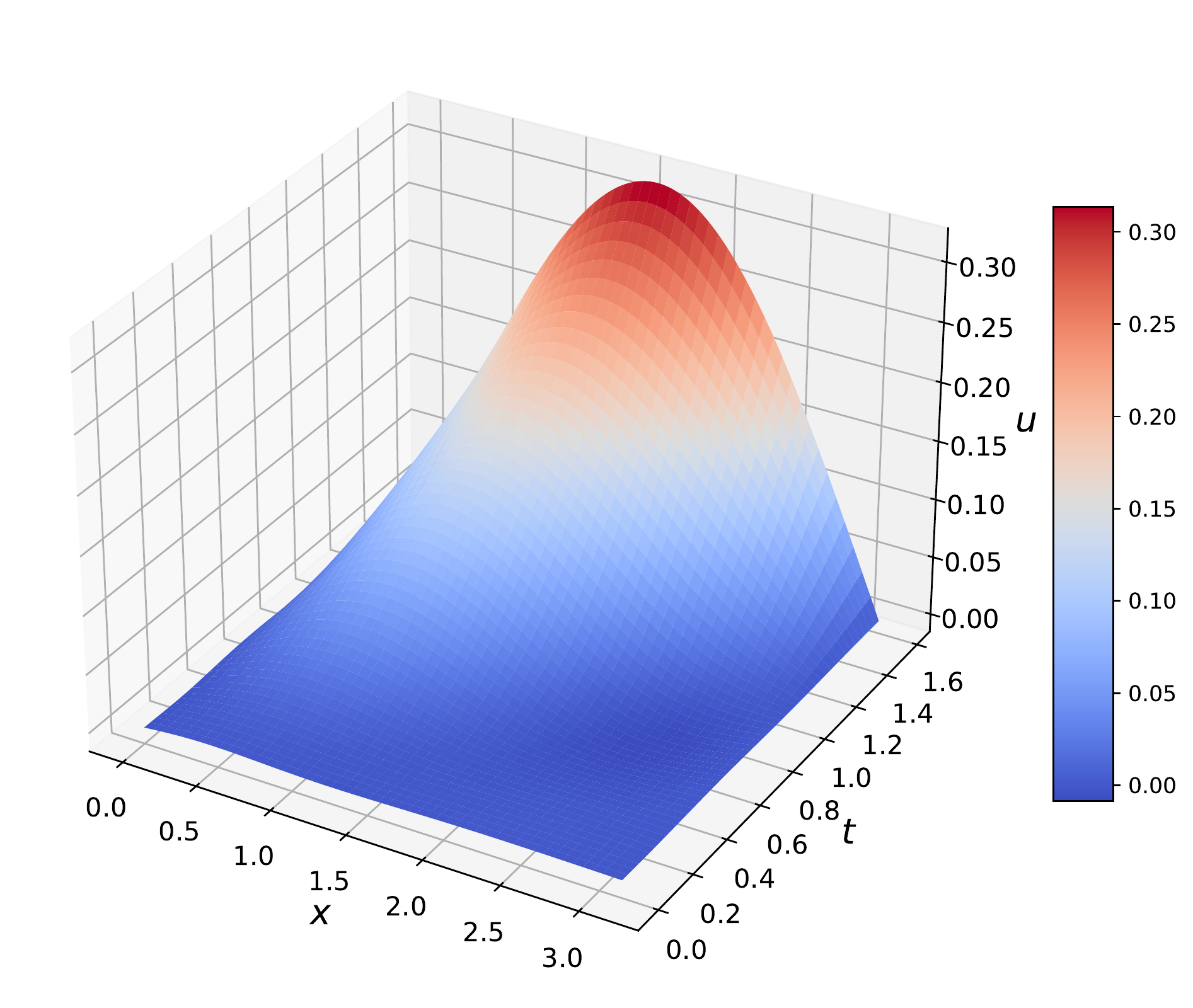}
\caption{\label{label7}Deflection profile of the beam for $p=1$ over the whole space-time domain after approximating the Dirac delta function by the Gaussian function.}
\end{minipage}\hspace{2pc}%
\begin{minipage}{18pc}
\includegraphics[width=18pc]{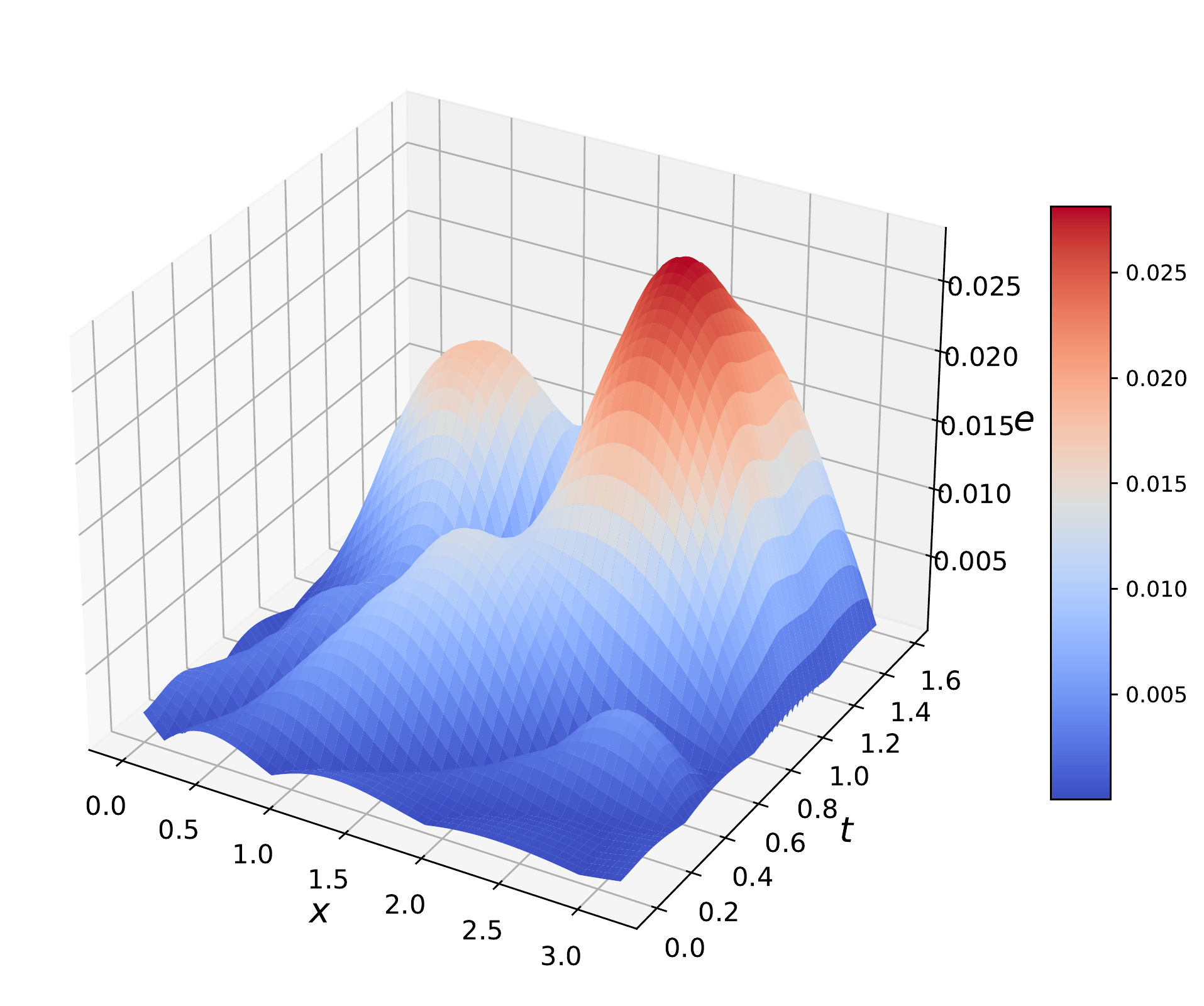}
\caption{\label{label8}The absolute error of the beam for $p=1$ over the whole space-time domain when the Dirac delta function is approximated using Gaussian function., \\ $e = | u_\mathrm{pred} - u_\mathrm{exact}|$.}
\end{minipage} 
\end{figure}

\begin{table}[ht]
\caption{\label{tab1}Relative error percent $R$ for mid-time deflections for different loads $(p)$.}
\begin{center}
\begin{tabular}{llll}
\br
$p=1$&$p=2$&$p=4$\\
\mr
0.072510$\%$ & 0.089523$\%$ & 0.079404$\%$\\
\br
\end{tabular}
\end{center}
\end{table}

As shown in Figure 6, the deflection predictions at the final time for different loads are accurately approximated using PINNs. The surface plot of deflection over the whole space-time domain is given in Figure 7. Figure 8 shows the absolute error plot. 

\subsection{PINNs for the inverse problem of moving load with Gaussian function}
\begin{figure}[ht]
\centering
\includegraphics[width=20pc]{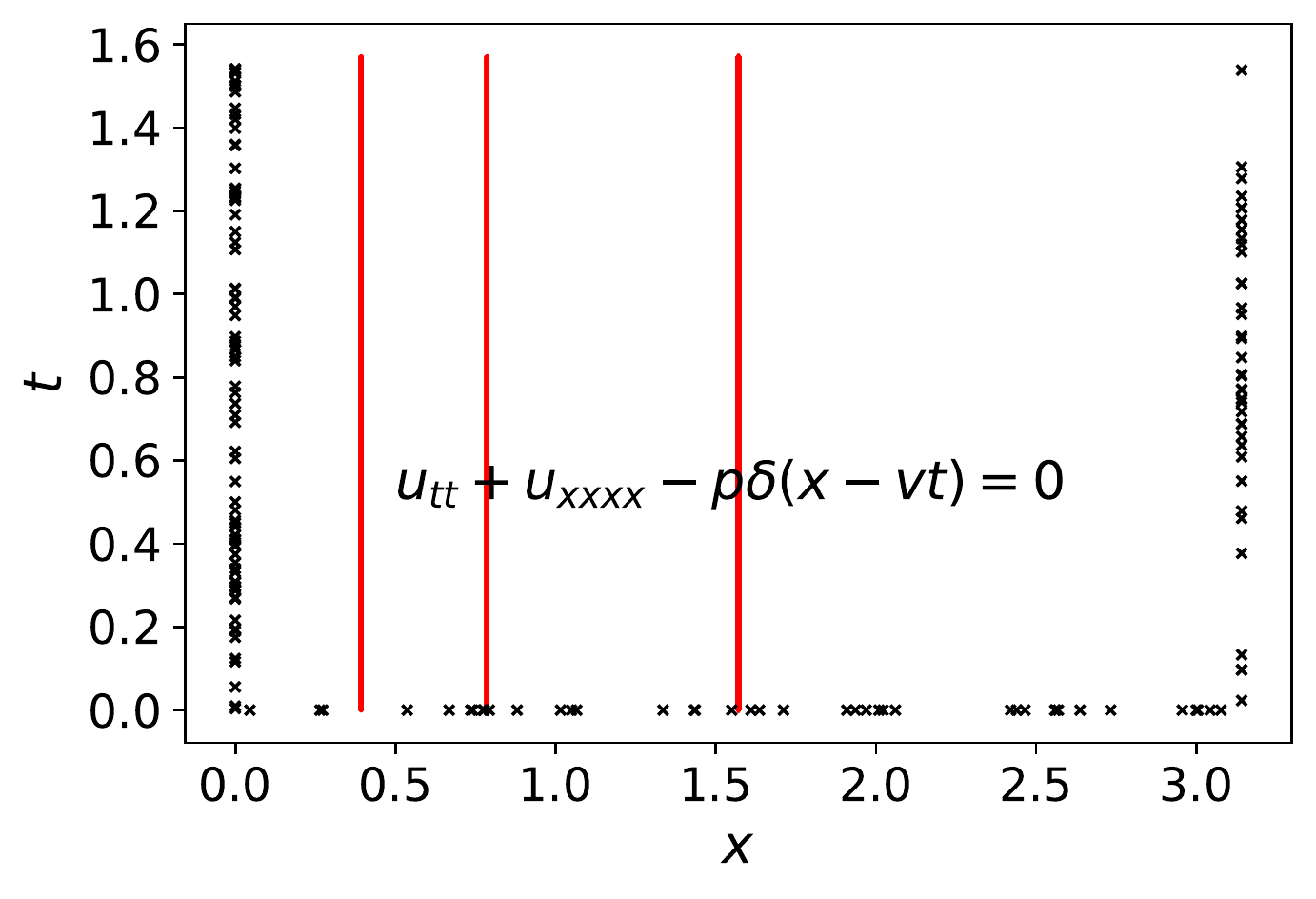}\hspace{2pc}%
\caption{\label{label9}Schematic plot of the data for the inverse problem: For the interior domain, ill-posed PDE is provided with $p$ as unknown \textbf{Red dots} Additional data points provided at spatial locations $x =\pi/8, \pi/4, \pi/2 $. \textbf{Black dots} Initial and boundary points.}
\end{figure}
For solving the inverse problem of moving load, a quantity of interest is $p$, which is the magnitude of the load. The magnitude of the load is a critical factor in determining the response of the structure. Greater values of $p$ result in a higher impact on the beam system, leading to increased displacement and stress. As a result, when analyzing the response of a structure to a moving load, it is important to consider the magnitude of the load. Additionally, it is crucial to consider the magnitude of the load when designing and evaluating structures that will be exposed to moving loads. 

The prior physical knowledge of moving load is ill-posed because an unknown quantity in the PDE is $p$. We need additional data for training the neural networks to predict $p$. The additional data of deflection of the beam is provided for 5000 points at three different locations at $x = \pi/8, \pi/4$ and $t = \pi/2$ as shown in Figure 9. This data can be collected using sensors. In this case, we provided the deflection from the forward PIML predictions. For training, the neural network architecture has four hidden layers, 20 neurons, and weights = 1 are considered. Using the tanh activation function, 2500 epochs were performed. The load magnitude prediction for $p = 1$, and $p = 4$ are $1.1343$ and $4.0668$, respectively.

\section{Conclusions and future work}
We introduced PIML for solving forward and inverse moving load problems. This approach involves using PIML to predict the deflection and magnitude of a load for a moving load problem on a simply supported beam. However, we found that vanilla PINNs struggle to train effectively while explicitly using a Dirac delta function. To overcome this challenge, we modified the physical model of the moving load with an approximation of the Dirac delta function with a Gaussian function in the PINN architecture. This modification allowed us to achieve accurate deflection predictions under different magnitudes of loading conditions. Traditional approaches for solving an inverse problem are iterative and computationally expensive because of many forward simulations. However, PINNs for inverse problems are comparatively computationally less expensive because PINNs do not need to solve the forward problems iteratively.

For the forward problems of moving loads, we observed that PINNs are not accurately predicting the solution for lower time levels because of inductive bias, implying that neural network training for higher time levels is better than for lower ones. In our future work, we aim to tackle this challenge by employing causal physics-informed neural networks. Additionally, different boundary conditions, loading conditions, velocities and noisy data are interesting to study. Tackling the Dirac delta function using the Fourier transform or integrating the physical equation is also a possible future research direction.
\section*{References}
\bibliographystyle{iopart-num}
\bibliography{ref}

\end{document}